# Cognitive Approach to Hierarchical Task Selection for Human-Robot Interaction in Dynamic Environments

Syed Tanweer Shah Bukhari[1,2], Bashira Akter Anima[3], David Feil-Seifer[3] and Wajahat Mahmood Qazi[2]

*Abstract—* In an efficient and flexible human-robot collaborative work environment, a robot team member must be able to recognize both explicit requests and implied actions from human users. Identifying "what to do" in such cases requires an agent to have the ability to construct associations between objects, their actions, and the effect of actions on the environment. In this regard, semantic memory is being introduced to understand the explicit cues and their relationships with available objects and required skills to make "tea" and "sandwich". We have extended our previous hierarchical robot control architecture to add the capability to execute the most appropriate task based on both feedback from the user and the environmental context. To validate this system, two types of skills were implemented in the hierarchical task tree: 1) Tea making skills and 2) Sandwich making skills. During the conversation between the robot and the human, the robot was able to determine the hidden context using ontology and began to act accordingly. For instance, if the person says "I am thirsty" or "It is cold outside" the robot will start to perform the tea-making skill. In contrast, if the person says, "I am hungry" or "I need something to eat", the robot will make the sandwich. A humanoid robot Baxter was used for this experiment. We tested three scenarios with objects at different positions on the table for each skill. We observed that in all cases, the robot used only objects that were relevant to the skill.

## I. INTRODUCTION

Recent developments in the creation of intelligent robots have created possibilities for collaborative work between people and robots in dynamic settings [1], [2]. As a result, it becomes more important for robots and other agents to comprehend teammates' implicit and explicit cues and translate those cues into suitable actions [1], [3]. If we can tell our teammate (a human) that "It is getting cold outside" or "I am feeling thirsty" rather than "I want to drink cold tea using a yellow cup," and in other situations, "I am hungry" or "I need something to eat" rather than "I want to eat burger placed at the right side," we can demonstrate the importance of understanding the environment. The teammate will infer the connection between "cold weather", "thirst" and "drink, and "hunger", "food" and "eat". The relationship between the two is that "cold weather" induces "thirst" and a desire to "drink," whereas "hunger" elicits a desire to "eat" or "consume" something. A colleague will therefore offer something to "drink" and another person will offer something to "eat" as a result.

A robot would be expected to behave similarly to a human teammate when collaborating in a team with a person [1] [4] [5] [6]. Although there have been several contributions in this area, this kind of cooperation is still difficult in human-robot interaction [1], [7], [8], [9]. Semantic association between words, items, and abilities can be a useful method to understand partial or incomplete information. The human-robot interaction (HRI) experience can be enhanced by the robot's capability to determine what the user wishes it to do next based on a hazy or imprecise command given a knowledge model of the activities and objects in the environment [1].

To meet these requirements, we have created a technique based on our current hierarchical control [3] and cognitive [1] designs that enables people and machines to collaborate on activities like making tea, sandwiches, burger, coffee, etc. together. In this respect, we have taken into account cognitive modalities such as actuators (Robot: Baxter, verbal response), working memory (semantic analysis, *Moveit* module, and *Rasa* chatbot), semantic memory, perception (lingual and verbal), and sensory memory.

A robot is needed to track activities, understand the commands and cues of teammates, and execute the required task(s) [10] [3]. In the past, researchers have looked at task coordination to motivate users to complete various subtasks carried out by a robot [3], communicate about task failures [7], and create new tasks from vocal instructions [10]. If people and machines can communicate vocally to discuss how to carry out challenging jobs, it will resemble a human-human interaction approach. However, such interaction has the added difficulty of teammates communicating with incomplete information or requests that leverage the knowledge of the task and the environment.

In this study, we present a system where the robot can complete the intended job by selecting hierarchical sub-tasks stored in procedural memory and can grasp the context of the environment in working memory using semantic similarity, RASA-based natural language understanding (NLU) engine. For task execution in a dynamic environment based on perceptual and semantic connections, we used a cognitive architecture (see Fig. 1). We use three scenarios to test our research, positioning the task items for various talents at various positions in front of the robot. By speaking with the human in each situation, the robot may use ontology to comprehend the context of the environment. The robot selects the skill it needs to do based on the semantic similarity score, performs the skill following the hierarchical task design, and uses the objects that come within the performing skill. The knowledge representation (KR) based on the existing study [1] is further tailored to accommodate the development of new skills in robot training/teaching mode [1].

*This work was supported by the National Science Foundation and IEEE Robotics & Automation Society (RAS) under 2021 Developing Country Faculty Engagement Program.

[1]Cognitive Robotics Group, Faculty of IT and CS, University of Central Punjab, Lahore, Pakistan (corresponding author; e-mail: stsbukhari@gmail.com , tanweer.shah@ucp.edu.pk )

[2]Intelligent Machines & Robotics Group, Department of Computer Science, COMSATS University Islamabad, Lahore Campus, Pakistan (email: wmqazi@cuilahore.edu.pk)

[3]Socially Assistive Robotics Group, Department of Computer Science & Engineering, University of Nevada, Reno, Reno, NV 89557, USA (e-mail: banima@nevada.unr.edu , dave@cse.unr.edu )

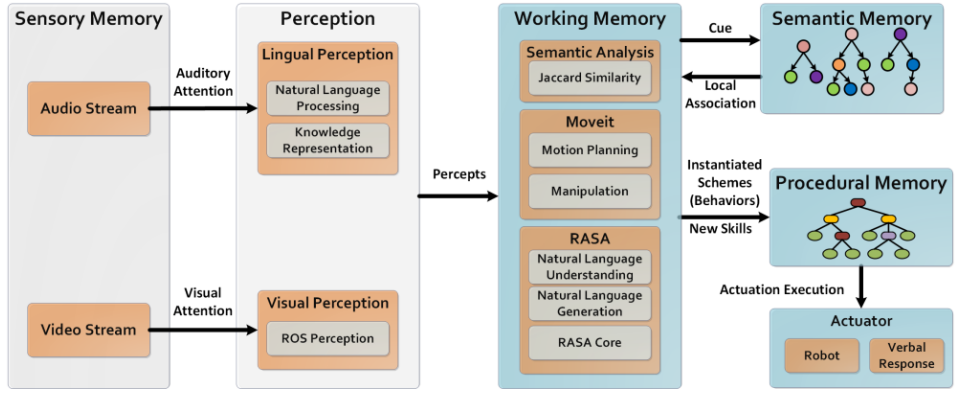

Fig 1 - NiHA's Modified Cognitive Architecture with Upgraded Perception Layer, Working Memory, and Procedural Memory [1]

## II. RELATED WORK

In human-robot interaction (HRI), robots with similar task representations can show effective results in collaborating with human teammates [5]. In a heterogeneous environment, communication is likely necessary for successful cooperation between robot and human teammates cooperation to complete tasks [8]. The clear sharing of information might serve as the foundation for communication. For instance, if the robot is instructed to "choose me a red bottle," it will be able to examine its surroundings, look for the object, and do the necessary actions to resolve the issue [1]. Tasks like route planning [11], human navigation guidance [12], learning [13], and task execution [4] may all be taught or created using explicit signals. In a related contribution, a vocal command-based interactive method was used to let people teach tasks to a mobile service robot [13]. Nicolescu explored how robots may learn tasks from language-based commands and advanced a creative strategy [10]. For socially conscious navigation in public settings, context identification, object detection, and scene data were used to generate context-appropriate rules [14]. It is necessary to create linkages between items, their effects, and the actions performed by robots to comprehend their environment and verbal signals from a human teammate [15]. Anthologies have been employed in addition to verbal cues to establish a connection between objects and their attributes [1], [16], [17]. Although this slightly enhanced the HRI experience but only a few relationship types—namely, isA, hasA, prop, usedFor, on, linked-to, and homonym—were able to extract information from implicit signals [16]. Ontology in the form of semantic memory was also described [17], [18], but it was unable to analyze the scenarios like "I'm feeling hungry," in which the robot understands the necessity to make the sandwich.

For interpreting explicit cues, we have developed semantic memory from WordNet and ConceptNet. This memory is further utilized for a similarity score between verbal cues, readily available objects (teapot, lettuce, meat, bread, etc.) on the table, and skills learned by a robot (i.e. tea and sandwich making). As a baseline control structure, we adopted a modified version of Nature-inspired Humanoid Cognitive Architecture for Self-awareness and Consciousness (NiHA) [1] (see Fig 1) and hierarchical control architecture [7], [3]. as part of procedural memory. Our previous hierarchical architecture [7], [3] involved humans and robots executing the entire tree to accomplish a specific task. We have revised hierarchical architecture [7], [3] to accommodate the learning of new skills using the knowledge representation module. Upon receiving the highest similarity score among the available task objects, the architecture performs the skill associated with that object.

## III. METHODOLOGY

### A. Sensory Memory

Sensory memory is part of short-term memory, which is further classified into iconic and echoic memory. The iconic involves the processing of brief images from a video stream whereas the echoic memory processes auditory steam.

### B. Perception Layer

*1) Lingual Perception*

The lingual perception has two categories, first is based on the Natural Language Processing (NLP) layer which is further composed of the Part-Of-Speech (POS) tagger, and Tokenization module Tokenization modules tokenize the spoken commands into words as nouns, verbs, and adjectives. The second part is based on a knowledge representation module specially tailored from existing work [1] to generalize the procedural memory to accommodate new skills in the form of recipes. It contains SkillNode, ObjectNode, ActionEdge(PicknPlace), SequentialNode(THEN), NonOrderingNode(AND), and AlternativePathNode(OR). Further details related to POS tags represented in words (nouns, verbs, and adjectives) can be accessed at [19].

*2) Visual Perception:*

The visual perception module can be developed with various deep learning modules. To simplify the process and to test various robot skills and cognitive capabilities we have opted for the ROS (Robot Operating System) defined Augmented Reality (AR) tags [20] to detect the objects on the table. AR

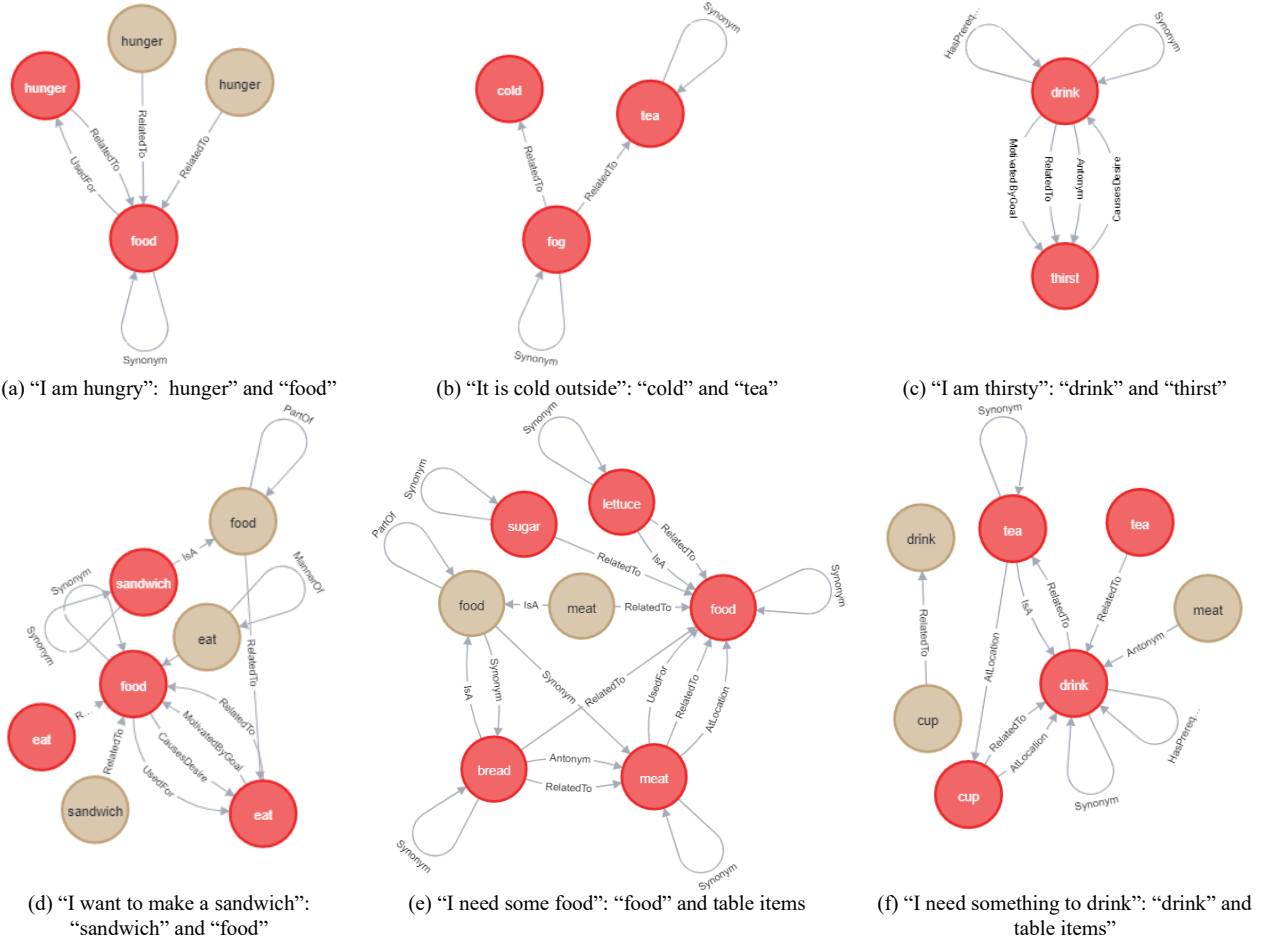

(a) "I am hungry": hunger" and "food"
(b) "It is cold outside": "cold" and "tea"
(c) "I am thirsty": "drink" and "thirst"
(d) "I want to make a sandwich": "sandwich" and "food"
(e) "I need some food": "food" and table items
(f) "I need something to drink": "drink" and table items

Fig 2 - Semantic Graphs extracted from Semantic Memory based on verbal cues.

tags help to identify and track the pose of the object to determine where the object is.

### C. Working Memory

Working Memory (WM) functions as an executive control that is aware of the current situation and can recall earlier events. The basic goal of WM is semantic processing, object grounding, motion planning, and motor command manipulations.

*1) Semantic Analysis*

The algorithm assesses the semantic similarity between spoken words and item categories present in the table-top scenario at the time. $Word_{Auditory} = \{word_1, word_2, word_3, ..., word_m\}$. The semantic function $\mathfrak{S}$ : $Word_{Auditory} \rightarrow Item$. The Similarity Index is being evaluated as

$$\mathfrak{S}(Word_{Auditory}, Item) = \frac{|Word_{Auditory} \cap Item|}{|Word_{Auditory} \cup Item|} \quad (1)$$

*2) Moveit*

To execute our experiment, Moveit [21] was used to plan and manipulate the robot's hand movement to perform pick and place objects from the surrounding environment.

*3) RASA Chatbot*

The RASA module has three components, natural language understanding (NLU), natural language generation (NLG), and RASA core. The RASA core acts executive control of the RASA environment. The NLU unit handles intent management whereas NLG is responsible for generating sentences based on predefined templates. We have used RASA as an intermediary between robot and human teammates.

### D. Semantic Memory

Semantic memory is developed from WordNet and ConceptNet having 117,659 Synsets (WordNet Nodes), 157,300 Lemma nodes, and 1653804 Concept (ConceptNet) nodes. There are 54 categories with 3730567 relationships [22]. Lemma nodes are the "root words" retrieved from the Concept nodes that can correlate Concept nodes completely or partially with Synsets whereas an assertion is considered an atom of knowledge in the Semantic Network [23]. The semantic memory is constructed as concept-relationship-feature or concept-relationship-concept i.e. Concept (Apple)-Relationship(IS_A)-Concept(Fruit) and Concept(Apple)-Relationship(isUsedFor)-Feature(Eating). Complete details about semantic memory can be accessed at [1]. Semantic memory is used during the processing of cues and the local association between available items and user commands (see Fig 2 for various examples).

### E. Procedural Memory

Procedural memory is what controls our actions and abilities. This recollection is wholly dependent on the kind of agent or robot being used. For the execution of skills, such as making tea and sandwiches, we have chosen Human-Robot Collaborative Architecture. The actions to be taken are detailed along with their hierarchical limitations by skills.

Table 1 - Semantic Similarity Score between Tagged Words (vertical) and Available Items (horizontal). This information is used to select which objects are most semantically related to words that the partner might say.

|          | Bread     | Cheese    | Cup       | Lettuce   | Meat      | Sugar     | Tea        | Teapot    |
|----------|-----------|-----------|-----------|-----------|-----------|-----------|------------|-----------|
| **Hot**      | 0.0080249 | 0.0043135 | 0.0049332 | 0.0023202 | 0.0069543 | 0.0065621 | **0.0116331** | 0.0011587 |
| **Hungry**   | 0.0006277 | 0.0000000 | 0.0000000 | 0.0000000 | **0.0034459** | 0.0000000 | 0.0000000 | 0.0000000 |
| **Thirst**   | 0.0012563 | 0.0000000 | **0.0057803** | 0.0052356 | 0.0006873 | 0.0000000 | 0.0051282 | 0.0000000 |
| **Sandwich** | **0.0224423** | 0.0160966 | 0.0065459 | 0.0178571 | 0.0176075 | 0.0026882 | 0.0108120 | 0.0021978 |
| **Drink**    | 0.0100839 | 0.0046816 | **0.0175440** | 0.0032726 | 0.0074370 | 0.0154660 | 0.0146541 | 0.0013999 |
| **Food**     | **0.0287881** | 0.0090561 | 0.0068393 | 0.0048706 | 0.0253697 | 0.0132474 | 0.0080704 | 0.0002427 |
| **Burger**   | 0.0044108 | 0.0029791 | 0.0000000 | 0.0068027 | **0.0111111** | 0.0004470 | 0.0000000 | 0.0000000 |
| **Coffee**   | 0.0025157 | 0.0033104 | 0.0233112 | 0.0053050 | 0.0048309 | 0.0062926 | **0.0588923** | 0.0066401 |
| **Cold**     | 0.0055744 | 0.0054682 | 0.0035714 | 0.0026762 | 0.0061406 | 0.0026882 | **0.0115401** | 0.0000000 |

*1) Hierarchical Task Representation*

The hierarchical task architecture's goal is to make it possible for complicated tasks to be executed realistically by humans and robots. This task design is built on a complicated hierarchical task network that enables simultaneous human and robot work in the same environment. Nearly every single task in the real world can be divided into more manageable tasks and set up as a hierarchical task network. In the real world, the task can be made up of a set of sequential, non-sequential, and parallel sub-tasks.

Our robot control architecture lets the system encode tasks with different kinds of constraints, such as SequentialNode(THEN), NonOrderingNode(AND), and AlternativePathNode(OR) [24]. Tasks are shown in a tree structure ObjectNode and ActionEdge(PicknPlace). The tasks that need to be done on objects are shown by the ObjectNode, and the actions to be taken on objects are shown by the ActionEdge(PicknPlace).

For a task with so many tiers, each node in the architecture keeps track of a state made up of the following: 1) Activation Level: a number that shows how important its parent thinks it is to run and finish a certain node, 2) Activation Potential: a number that shows how useful the node is thought to be and is sent to the node's parent, 3) Active: a Boolean variable that says the behavior is active when the node's activation level is higher than a threshold. 4) Done: a Boolean variable that is set to true when the node has done its job. Each node always keeps track of the above state information. By doing both top-down and bottom-up spreading, the activation spreading technique makes sure that the task is done right based on the constraints.

To complete a task, activity-spreading messages are sent from the root node to its children to spread activity levels across the task tree. A bottom-up mechanism sends activation potential up the tree by having nodes send status messages to their parents about their current state. In each cycle, a loop helps keep the state of each node in the task structure up to date by checking the different parts of the node's state and adjusting them as required.

The controller architecture can handle more than one robot because it keeps a copy of the task tree for each robot. This includes when that robot is currently working on behavior when a robot has completed one, and the activation potential and level for each robot and each behavior.

*a) Choosing Skill*

Here, first, it finds the object from the list which has the highest semantic score then finds which skill has this object. By doing this it finds out the skill that it wants to perform.

*Algorithm 1: Choosing Skill*

1: **For** object ∈ object_list **do**
2:    **If** object is highest_semantic_similarity_score **then**
3:       skill_object ← object
4:    **End If**
5: **End For**

6: **For** skill ∈ skill_list **do**
7:    **If** skill_object is in skill **then**
8:       chosen_skill ← skill
9:    **End If**
10: **End For**

After choosing the skill the hierarchical architecture updates its activation potential and activation level. For this, another behavior called skill_behavior was added to the previous hierarchical architecture [24] [25]. It chooses the skill it wants to execute from the hierarchical design it wants to execute. To do this it only spreads its Activation level value to the child nodes belonging to the chosen skill. This allows the child node with the *Skill* behavior to activate.

In the case of updating activation potential, the skill_behavior node spreads the activation potential of the single child with the chosen skill.

*Algorithm 2: Skill_behavior - Spread Activation*

1: msg ← {activation level = 1.0}
2: **For** child ∈ children **do**
3:    **If** child.skill is chosen_skill **then**
4:       SendToChild(child,msg)
5:    **End If**
6: **End For**

*Algorithm 3: Skill_behavior-update Activation Potential*

1: **For** child ∈ children **do**
2:    **If** child.skill is Chosen_skill **then**
3:       activation potential ← child.activation_potential
4:    **End if**
5: **End For**

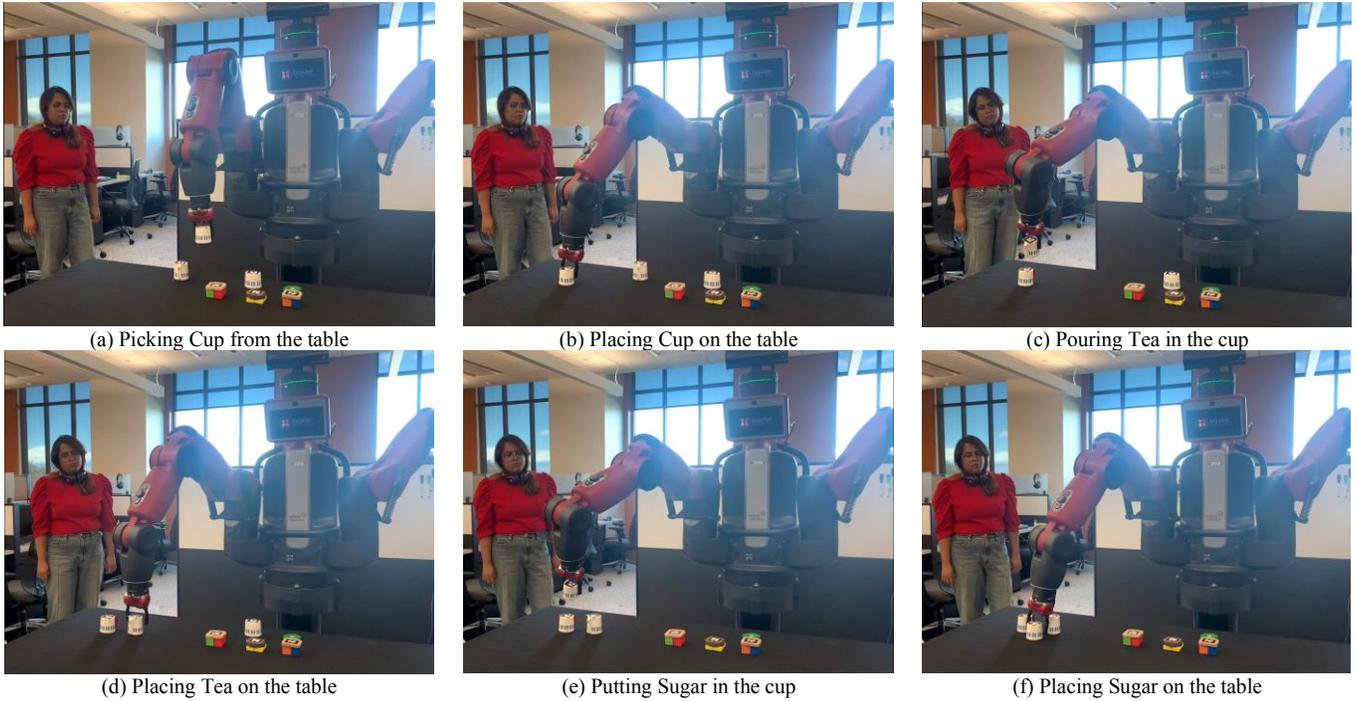

(a) Picking Cup from the table
(b) Placing Cup on the table
(c) Pouring Tea in the cup
(d) Placing Tea on the table
(e) Putting Sugar in the cup
(f) Placing Sugar on the table

Fig 3 - The robot is making a cup of tea after the human said, "It is cold outside." The robot determines to execute the Tea Making Skill after analyzing the semantic scores of the available table objects.

*2) Adding Skill Component in Hierarchical Architecture*

We added a new component to the prior task architecture to expand it. The skill that evaluates the surroundings and interactions to determine which of a variety of duties should be carried out. Following the contact between the person and the robot, the semantic knowledge module decides which task the robot should complete. The skill node receives a ROS message in string form; it can then decide which task needs to be executed. We can give the robot a variety of skill tasks under the SkillNode. Whenever the robot chooses a task to perform, it will perform the task accordingly. These skills are designed with Nodes like SkillNode(i.e. Tea and Sandwich making), THEN, AND, OR, ObjectNode, and ActionEdge(PicknPlace). As shown in Fig 3, there are two skills listed under the SkillNode: 1) Tea Making Skill and 2) Sandwich Making Skill. The Skill component determines which task to run based on semantic information and the objects that are available in the environment, the semantic relevance of various objects to words that a user might speak is shown in Table 1.

## IV. EXPERIMENT DESIGN

We created a speech conversation between a human participant and a robot to demonstrate the capabilities of the system we created and to verify the functionalities of the cognitive and hierarchical architecture. The robot can understand the hidden context and carry out a skill task using items from the nearby surroundings based on the participant's input. We experimented with a lab setting using a human participant and a Baxter humanoid robot that was positioned in front of a table with items. This experiment involves using a robot to make tea and sandwiches. A Kinect v2 camera on top of Baxter's head and Baxter's right-hand camera were used to detect the object's AR tags. The robot will decode the tagged word from the human's speech in this human-robot interaction and assess the items' semantic similarity scores (see Table I) about the decoded tagged word. The architecture will use the score to pick the most suitable skill task to execute. If the human says a statement like "I am thirsty" or "It is cold outside," the tagged words will be "thirsty" and "cold" respectively. Based on the similarity score, in both cases, it is observed that the objects under the TeaMaking Skill have the highest scores. As a result, the robot will decide to perform the TeaMaking Skill. Based on the task tree (see Fig 3), the task will be ((PicknPlace Cup) THEN ((PicknPlace Tea) AND (PicknPlace Sugar))). According to this task statement, the robot will first pick and place Cup, then pick and place Tea and Sugar in a non-ordered fashion (see Fig 4).

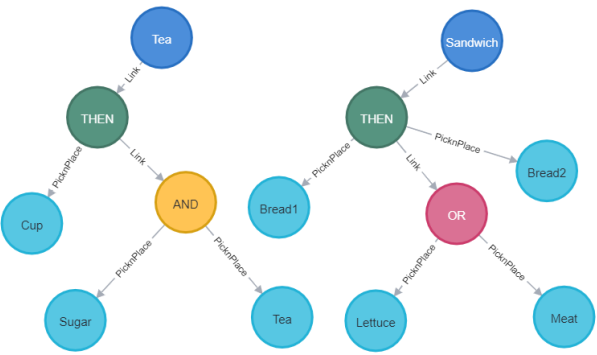

Fig 4 - A new component SkillNode was added to the hierarchical task tree which allows the system to choose the skill based on the similarity score. Two types of Skills: 1) Tea Making Skill and 2) Sandwich Making Skill were added under the SkillNode.

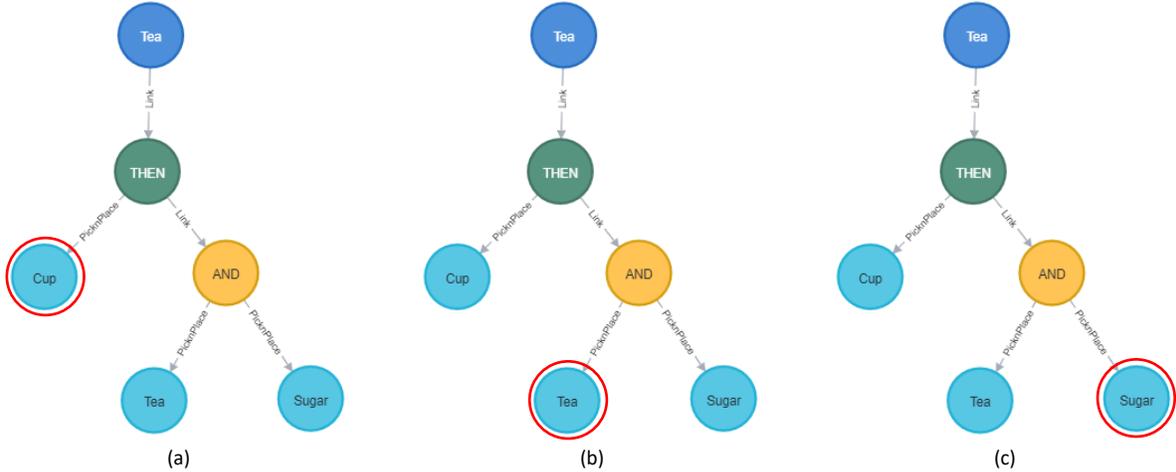

Fig 5 - Order of execution for Tea Making Skill - ((PicknPlace Cup)THEN((PicknPlace Tea)AND(PicknPlace Sugar))). (a) Tea Making Skill is invoked, which initiates the PickAndPlace process for the Cup object, (b) The PickAndPlace action for the Tea which makes the robot starts pouring Tea into the cup, (c) The PickAndPlace action for the Sugar under the AND node is activated which makes the robot adding sugar into the cup.

Fig. 5 displays the hierarchical state depiction of each stage involved in using the TeaMaking Skill.

In comparison, the Meat object from the object collection has the greatest semantic score connected to the tagged word "hungry" if the person states something like "I am hungry" (see Table 1). The robot will begin making a sandwich because Meat is represented by the SandwichMaking skill. The task will be ((PicknPlace Bread1)THEN((PicknPlace Meat)OR(PicknPlace Lettuce))THEN(PicknPlace Bread2), again based on the tree. Therefore, the robot would choose and put Bread1 before choosing and putting either Meat or Lettuce. The task will then be finished by the robot by picking up and placing Bread2.

## V. RESULTS

In our experiment, when the person says, "It is cold outside." Speech recognition provides the ontology with a word string spoken by the user. The Jaccard Similarity measure is used to determine the lexical similarity from the decoded speech between the labeled words and the accessible items, such as "tea," "sugar," "cup," "bread," "meat," "cheese," "lettuce," and "teapot" (see Table I). The ontology was able to identify the statement's inferred context based on the score, which shows that the spoken phrase is connected to "tea". The Tea Making task was chosen by the ontology using the list of objects that are both readily accessible and most closely related to the user's speech statement. This reflects a connection between the user's statement, the available objects on the table, and the available tasks that the robot can complete.

Fig 5 illustrates the step-by-step state for the tree nodes in our robot architecture for executing the Tea Making skill. In the first phase, when the skill node received the object name *"Tea"*, based on the highest similarity score between available items and *"Cold"* (see Fig2b for related graph). *"Tea"* has the highest semantic similarity score (*0.0115401* in Table I) i.e. 30.64%, among the other items i.e. Cup (9.48%), Sugar (7.14%), Lettuce (7.11%), Bread1 (14.80%), Bread2 (14.80%) and Meat (16.31%), the task tree then decided to execute the Tea Making Skill. The THEN node was activated for this skill (see Fig 5a), and the robot proceeded to pick and place the Cup (see Fig 4a Fig 4b respectively). When the robot placed the Cup on the table, the status of the Cup node was changed to Done from Active. From the task tree, the robot would activate the AND node (Fig 5b) and start picking the tea to pour into the cup (Fig 4c). After pouring the tea into the cup, the Tea was set on the table (Fig 4d), which made the Tea node in the task tree Done from Active. Then, the robot moved to the next step according to the task tree and activated the Sugar node (Fig 5c), and start to put sugar in the cup (Fig 4e). In the end, when the Sugar was placed on the table (Fig 4f), all the nodes' statuses were changed to Done, and the whole skill task was completed based on the tree design.

We had different table setups for experiments, but the robot was still able to figure out the concept of the surroundings and worked on the skill from the hierarchical task tree. Our observations indicated that the robot does not go for objects under different skill sets. Additionally, we provided two statements for each skill test to validate the case scenarios. For instance, we used statements like "I am thirsty" (see Fig2c for graph) and "It is cold outside" (see Fig2b for graph) for the Tea Making Skill. Likewise, for the Sandwich Making Skill, we used statements like "I am hungry" (see Fig2a for graph) and "I want to make a sandwich" (see Fig2d for graph). Furthermore, we have also tried queries "I need some food" (see Fig2e for graph) and "I need something to drink" (see Fig2f for graph), the respective similarity score about extracted action verbs, nouns, and adjectives can be found in Table I. Therefore, we can observe that based on the ontology approach, the system was able to understand the context behind the user statement "It is cold outside" and choose to perform a hierarchical skill task (Tea Making Skill) by identifying the relationship between the context and the objects nearby.

## VI. DISCUSSION AND FUTURE WORK

This paper proposes a way to offer an efficient and flexible human-robot collaboration environment in which the robot teammate can perform the user's desired task by deciphering both vague and clear requests in natural language form from a human teammate. The ontology played a vital role in the understanding of user commands due to the semantic relationship between various concepts and features. This architecture has the following contributions:

- The system can find an implied link between the context of the situation and the surrounding environment using the ontology approach after interacting with a human user.
- In our extended hierarchical task architecture, the robot will only select the hierarchical sub-tasks that are most relevant to the specific task derived from the ontology approach.

Currently, the robot is performing the skill task after interacting once with the human. However, in the future, we are planning to add more scope to hold a conversation to make the system more dynamic and diverse. Additionally, we are hoping to apply this ontology approach in a multi-human robot environment for more robust and diverse collaboration.

ACKNOWLEDGMENT

The authors would like to acknowledge the financial support of this work by the National Science Foundation (NSF #SES-2121387, #IIS-2150394). We acknowledge the funding support from IEEE Robotics & Automation Society (RAS) under 2021 Developing Country Faculty Engagement Program.